\documentclass[10pt,twocolumn,letterpaper]{article}
\usepackage{boldline}
\usepackage{cvpr}              

\usepackage[dvipsnames]{xcolor}

\definecolor{cvprblue}{rgb}{0.21,0.49,0.74}
\usepackage[pagebackref,breaklinks,colorlinks,citecolor=cvprblue]{hyperref}

\def\paperID{5583} 
\def\confName{CVPR}
\def\confYear{2024}
\usepackage{bbm}
\usepackage{amsmath,amsfonts}
\usepackage[ruled,vlined]{algorithm2e}
\usepackage{array}
\usepackage{textcomp}
\usepackage{stfloats}
\usepackage{url}
\usepackage{verbatim}
\usepackage{graphicx}
\usepackage{multirow}
\usepackage{booktabs}
\usepackage{comment}
\usepackage{float}
\usepackage{tabularx}
\usepackage{makecell}
\usepackage{arydshln}
\usepackage{times}
\usepackage{epsfig}
\usepackage{graphicx}
\usepackage{amsmath}
\usepackage{amssymb}
\usepackage[accsupp]{axessibility}

\definecolor{cvprblue}{rgb}{0.21,0.49,0.74}
\usepackage[pagebackref,breaklinks,colorlinks,citecolor=cvprblue]{hyperref}
\usepackage{listings}
\DeclareCaptionStyle{ruled}{labelfont=normalfont,labelsep=colon,strut=off} 
\floatstyle{ruled}
\newfloat{listing}{tb}{lst}{}
\floatname{listing}{Listing}
\def\paperID{7119} 
\def\confName{CVPR}
\def\confYear{2024}

\title{Causal Mode Multiplexer: A Novel Framework for Unbiased Multispectral Pedestrian Detection}
\author{
    Taeheon Kim\thanks{Equally contributed.} \! $^{1}$, 
    Sebin Shin\footnotemark[1] \! $^{1}$,
    Youngjoon Yu$^{1}$,
    Hak Gu Kim$^{2}$,
    Yong Man Ro\thanks{Corresponding author.} \! $^{1}$\\
    $^{1}$Integrated Vision and Language Lab, KAIST, South Korea\\
    $^{2}$Immersive Reality and Intelligent Systems Lab, Chung-Ang University, South Korea
    \\
    \{eetaekim, ssbin0914, greatday, ymro\}@kaist.ac.kr, hakgukim@cau.ac.kr
}

\begin{document}
\maketitle

\begin{abstract}
RGBT multispectral pedestrian detection has emerged as a promising solution for safety-critical applications that require day/night operations. However, the modality bias problem remains unsolved as multispectral pedestrian detectors learn the statistical bias in datasets. Specifically, datasets in multispectral pedestrian detection mainly distribute between ROTO\footnote{R$\star$T$\star$ refers to the visibility (O/X) in each modality. Generally, ROTO refers to daytime images, and RXTO refers to nighttime images. ROTX refers to daytime images in obscured situations.} (day) and RXTO (night) data; the majority of the pedestrian labels statistically co-occur with their thermal features. As a result, multispectral pedestrian detectors show poor generalization ability on examples beyond this statistical correlation, such as ROTX data. To address this problem, we propose a novel Causal Mode Multiplexer (CMM) framework that effectively learns the causalities between multispectral inputs and predictions. Moreover, we construct a new dataset (ROTX-MP) to evaluate modality bias in multispectral pedestrian detection. ROTX-MP mainly includes ROTX examples not presented in previous datasets. Extensive experiments demonstrate that our proposed CMM framework generalizes well on existing datasets (KAIST, CVC-14, FLIR) and the new ROTX-MP. Our code and dataset are available at: \href{https://github.com/ssbin0914/Causal-Mode-Multiplexer.git}{https://github.com/ssbin0914/Causal-Mode-Multiplexer.git}.
\end{abstract}
\vspace{-0.3cm}
\section{Introduction}
\noindent \indent Multispectral pedestrian detection plays a critical role in many real-world applications that require both day/night operations, such as smart surveillance cameras (CCTVs), search and rescue (SAR) autopilots, and autonomous vehicles (AVs)~\cite{kim2022defending, hwang2015multispectral,choi2018kaist, c:25,gonzalez2016pedestrian, 9419080, cao2021handcrafted, qingyun2022crossmodality, zhou2020improving, kim2022map, zhang2019weakly,kim2024mscotdet}. Despite its notable progress, an overlooked factor in multispectral pedestrian detection is the modality bias problem that occurs in multimodal models (e.g., Visual Question Answering) ~\cite{niu2021counterfactual, cadene2019rubi,zhang2023layout, kv2020reducing, wen2021debiased}. Multimodal models are known to often leverage spurious correlations between a certain modality and answers due to statistical biases in datasets~\cite{yang2021causal, zang2023discovering, nan2021interventional, liu2022show, huang2022deconfounded}. For example, a Visual Question Answering (VQA) model may blindly answer “tennis” for the question “What sports?” just referring to the most co-occurring textual QA pairs (i.e., linguistic bias) in the train data~\cite{ramakrishnan2018overcoming, niu2021counterfactual}. Models exploiting statistical bias in datasets often demonstrate poor generalization ability to out-of-distribution data, rarely providing proper multimodal evidence for prediction~\cite{wu2022characterizing, gat2020removing}. \\
\begin{figure}[t]
  \centering
  \includegraphics[width=0.85\linewidth]{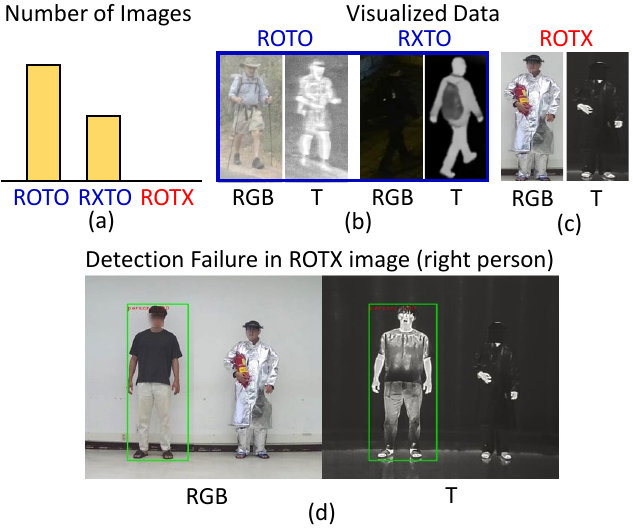}
  \caption{(a) ROTO, RXTO, and ROTX distribution in existing multispectral pedestrian datasets. (b) \textcolor{blue}{In the existing datasets}, a high statistical correlation exists between pedestrian labels and their thermal features. (c) Example of a ROTX image. (d) Current multispectral pedestrian detection models fail prediction in \textcolor{red}{ROTX} image (the right person) due to learning statistical bias in datasets.}
    \label{fig:1}
    \vspace{-0.5cm}
\end{figure}
\indent As modality biases can significantly impact the reliability of multimodal models, it is essential to investigate statistical bias in multispectral pedestrian datasets~\cite{hwang2015multispectral, gonzalez2016pedestrian, c:25, jia2021llvip}. Analyzing their distribution, we find an ever-overlooked statistical bias. Illustrated in Fig.\ref{fig:1} (a), most data are distributed between ROTO (daytime images) or RXTO (nighttime images). In these data, pedestrian labels always statistically co-occur with their thermal features, as thermal sensors can generally capture robust pedestrian silhouettes all day/night. On the other hand, there are rarely ROTX images that can be captured in thermal obscured situations in the daytime e.g., firefighters wearing heat-insulation cloth as in Fig.\ref{fig:1} (c). From this data distribution, we hypothesize that models may learn the statistical co-occurrence between the pedestrian label and the presence of thermal features. If so, there is a potential risk that multispectral pedestrian detectors may fail in scenes in which such statistical correlation does not hold, e.g., ROTX images in which thermal features are obscured.\\
\indent In this paper, we test multispectral pedestrian detection models under ROTX data and validate our hypothesis. Since existing datasets~\cite{hwang2015multispectral, gonzalez2016pedestrian, c:25, jia2021llvip} do not include ROTX data, we first collect ROTX pedestrian images based on practical scenarios. This includes search and rescue (SAR) on firefighters and surveillance of pedestrians over a window. In this situation, pedestrians are visible in only RGB but obscured in thermal as the intermediate obstacle (i.e., heat-insulating cloth or window) blocks the thermal radiation emitted by the pedestrian from reaching the thermal sensors. Fig.\ref{fig:1} (d) demonstrates that a multispectral pedestrian detection model fails prediction on ROTX, though the firefighter is obviously visible in RGB. As seen, multispectral pedestrian detectors make predictions on ROTX just based on the absence of the thermal feature, motivating us to formulate this phenomenon as a modality bias problem.\\
\indent To address this problem, we propose a novel Causal Mode Multiplexer (CMM) framework that performs unbiased inference from statistically biased multispectral pedestrian training data. We adopt this problem setting instead of explicitly changing the training priors (e.g., adding ROTX in the train set) to verify that multispectral pedestrian detectors can disentangle the learned multimodal knowledge and memorized priors by learning \textit{causality}. Specifically, the CMM framework learns causality based on different cause-and-effects between ROTO, RXTO, and ROTX inputs and predictions. For ROTO data, we guide the model to learn the \textit{total effect}~\citep{pearl2018book} in the common mode learning scheme. Next, for ROTX and RXTO data, we utilize the tools of counterfactual intervention to eliminate the direct effect of thermal by subtracting it from the total effect. To this end, we modify the training objective from maximizing the posterior probability likelihood to maximizing the \textit{total indirect effect}~\citep{pearl2018book} in the differential mode learning scheme. Our design requires combining two different learning schemes; therefore, we propose a Causal Mode Multiplexing (CMM) Loss to optimize the interchange.\\ 
\indent Moreover, we evaluate the modality bias in multispectral pedestrian detection with our new dataset: ROTX Multispectral Pedestrian (ROTX-MP) dataset. Different from existing multispectral pedestrian datasets~\cite{hwang2015multispectral, gonzalez2016pedestrian, c:25} which are constrained to ROTO (day) and RXTO (night) data, ROTX-MP mainly includes ROTX pedestrians comprised of 1000 test image pairs. Our experimental results demonstrate that our CMM framework generalizes well under ROTX-MP even with biased training and also performs robustly on existing datasets~\cite{hwang2015multispectral, c:25, gonzalez2016pedestrian}. \\
The main contributions of our paper are:
\begin{enumerate}
  \item We propose a Causal Mode Multiplexer (CMM) framework that learns different causality between ROTO, RXTO, and ROTX inputs and pedestrian labels in multispectral pedestrian detection.
  \item We propose a Causal Mode Multiplexing (CMM) Loss to optimize the interchange between learning different causal representations.
  \item To evaluate modality bias in multispectral pedestrian detection, we conduct a new dataset: ROTX-MP. 
  \item Extensive experiments demonstrate that our CMM framework generalizes well under ROTX test data with even biased training data - ROTO, RXTO.
\end{enumerate}
\section{Preliminaries}
\noindent \indent Before introducing our method, we present the fundamental concepts of causal inference~\citep{pearl2000models, pearl2018book, robins2003semantics, pearl2016causal, bareinboim2012controlling, pearl1995causal, rubin2005causal}.
\subsection{Structural Causal Model (SCM)} 
\noindent \indent Structural Causal Models reflect the cause-and-effect relationships (\textit{links} $\mathcal{E}$) between the set of variables (\textit{nodes} $\mathcal{V}$). The cause-and-effects (\textit{cause} $\rightarrow$ \textit{effect}) are represented in an acyclic graph $\mathcal{G}$= $\{\mathcal{V}, \mathcal{E}\}$. For random variables $X$ and $Y$ which the direct effect of $X$ is on $Y$, the direct link could be formulated as $X \rightarrow  Y$. If an indirect effect of $X$ is on $Y$ through the variable $M$, $M$ is considered a mediator between $X$ and $Y$ ($X\rightarrow M \rightarrow Y$ in Fig.\ref{fig:2} (a)). With structural causal models, the examination of causality links among variables can be achieved through variable intervention, which involves modifying the value of particular variables and subsequently observing the outcomes. 
\subsection{Counterfactual Intervention}
\begin{figure}[t]
  \centering
  \includegraphics[width=0.8\linewidth]{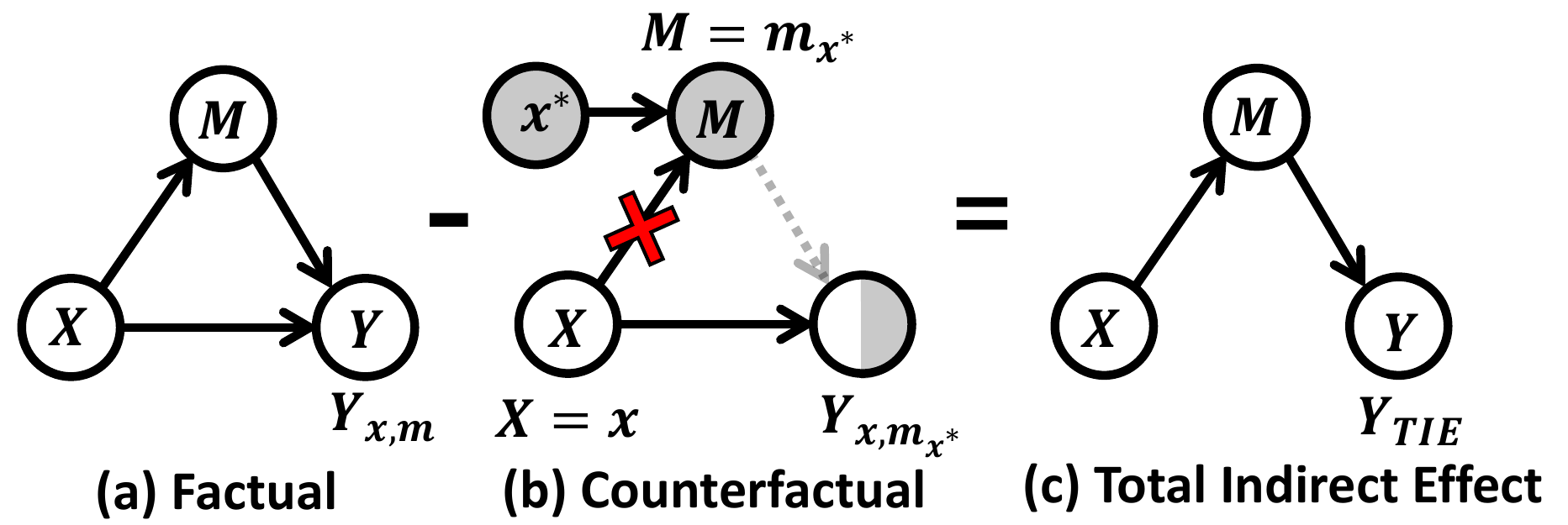}
  \caption{Structural Causal Models (SCMs) of (a) factual, (b) counterfactual, and (c) total indirect effect scenarios. The direct effect of $X \rightarrow Y$ can be eliminated due to counterfactual intervention.}
    \label{fig:2}
    
\end{figure}

\noindent \indent Counterfactual intervention can break the direct link and eliminate the effect of particular variables. To do this, the effects on $Y$ are compared from two different treatments to the cause variable $X$: factual and counterfactual. Take the SCM in Fig.\ref{fig:2} (a) as an example i.e., $X\rightarrow M \rightarrow Y$. Suppose that $X=x$ represents the ``treatment condition” and $X=x*$ represents the ``no-treatment condition" (lowercase letter indicates the observed value of the random variable). Then we can consider factual and counterfactual scenarios. \\
\hspace{-0.25cm}\textbf{Counterfactual Notations}. We denote the value that $Y$ would obtain in the factual scenario if $X$ is assigned $x$ and $M$ is assigned $m$ as $Y_{x,m_{x}}=Y(X=x, M=m_{x})$. Similarly, in the counterfactual scenario, $Y$ would have the value $Y_{x,m_{x*}}=Y(X=x, M=m_{x*})$ when $X=x$, $M=m_{x*}=M(X=x*)$. $X$ is set to a value $x*$ (usually zero value or mean features) and can break the direct link between $M$ and its parent node $X$. We note that in the counterfactual scenario, $X$ can be simultaneously assigned to different values $x$ and
$x*$. So when the intervention is conducted on $M$, the variable $X$ retains its original value of $x$ as if $x$ (observation on $X$) had existed as in Fig.\ref{fig:2} (b).\\
\textbf{Total Indirect Effect}. Now we can estimate the total indirect effect (TIE) by comparing two hypothetical scenarios. \\
\begin{equation}{
TIE=Y_{x,m_{x}}-Y_{x,m_{x*}}.
\label{eq:1}
}\end{equation}
\noindent Total indirect effect (TIE), as in Fig.\ref{fig:2} (c) breaks the direct link of $X \rightarrow Y$. Furthermore, total indirect effect (TIE) can be decomposed into total effect (TE) and natural direct effect (NDE). 
\begin{equation}{
\begin{aligned}
TIE&=TE-NDE \\
&=(Y_{x,m_{x}}-Y_{x*,m_{x*}})-(Y_{x,m_{x*}}-Y_{x*,m_{x*}}).
\end{aligned}
\label{eq:2}
}\end{equation}
When we examine the total effect (TE), we compare two hypothetical scenarios: one where $X = x$ and the other where $X = x*$.  In contrast, the natural direct effect (NDE) represents the effect of $X$ on $Y$ while keeping the mediator $M$ blocked. It measures the change in $Y$ as $X$ transitions from $x*$ to $x$, with $M$ assigned to the value of the no-treatment $X = x*$, thereby nullifying $M$'s response to the treatment $X = x$. In the subsequent section, we will take a more in-depth look at the interpretations of these effects in the context of multispectral pedestrian detection.
\section{Structural Causal Model of Multispectral Pedestrian Detection}
\noindent \indent Before pruning direct effects and performing counterfactual interventions, we investigate the causal links (causality) in multispectral pedestrian detection as below:\\
\begin{figure}[t]
  \centering
  \includegraphics[width=0.7\linewidth]{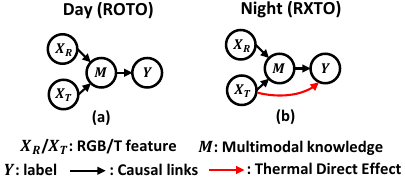}
  \caption{Structural Causal Models (SCMs) of multispectral pedestrian detection on (a) ROTO and (b) RXTO data. (b) The thermal direct effect is considered in RXTO scenarios as models rely heavily on the thermal features for making predictions in the nighttime.}
    \label{fig:3}
\end{figure}
\hspace{-0.15cm}\noindent\textbf{Link\,$\mathbf{X_{R}\rightarrow M}\leftarrow\mathbf{X_{T}}$ (Feature Fusion)}. RGB and thermal features ($\mathbf{X_{R}}$ and $\mathbf{X_{T}}$) are fused to generate fusion features of multimodal knowledge $M$. \\
\textbf{Link\,$\mathbf{M\rightarrow Y}$ (Class Label Prediction with Multimodal Knowledge)}. Using the multimodal knowledge $M$, the head network outputs class prediction $Y$ for each fused feature.\\
\indent Beyond the above causal links, an additional direct link should be considered in practice. In RXTO data (nighttime), the predictions heavily rely on the thermal modality because RGB sensors degrade at night. Thus, a direct effect of the thermal feature $X_{T}$ on the prediction $Y$ can be formulated as a direct link below:\\
\textcolor{red}{\textbf{Link\,$\mathbf{X_{T}\rightarrow Y}$ (Thermal Direct Effect)}.} As machines memorize priors, models will learn the skewed preference toward thermal in RXTO data. This direct effect causes detection failures when inferring ROTX data. We formulate this cause-and-effect as the thermal direct link.\\
\indent From the above discussions, we represent structural causal models separately on (a) ROTO (daytime) data and (b) RXTO data (nighttime) in Fig.\ref{fig:3} (a)-(b). Our goal is to prune the direct link $\mathbf{X_{T}\rightarrow Y}$ when training RXTO.
\section{Proposed Method}
\subsection{Causal Mode Multiplexer}
\noindent \indent Based on the different causal graphs in ROTO and RXTO data, our key idea is to interchangeably learn causality from two different learning schemes. First, the 1) \textit{common mode} scheme is designed to learn the total effect in ROTO data (daytime data). Second, the 2) \textit{differential mode} learning scheme learns the total indirect effect in RXTO data (nighttime images) to prune the thermal direct effect.\\
\begin{figure*}[t]
  \centering
  \includegraphics[width=0.9\linewidth]{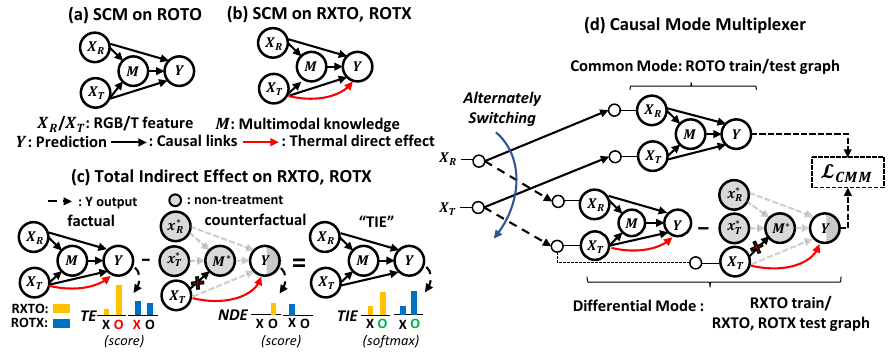}
  \caption{Our Structural Causal Model (SCM) formulations on (a) ROTO (day) and (b) RXTO (night) data. (a) We add direct links ($X_{R} \rightarrow Y$, $X_{T} \rightarrow Y$) to the conventional causal graph to determine modes and measure the thermal direct link. (b) The thermal direct link (red) hinders the model from learning causality. (c) implies the total indirect effect on RXTO and ROTX for which the thermal direct link is pruned. (d) We propose a Causal Mode Multiplexer framework that learns causality both on ROTO and RXTO data and thus generalizes well on all ROTO, RXTO, and ROTX.}
    \label{fig:4}
\end{figure*}
\hspace{-0.14cm}\noindent\textbf{1. Common Mode - ROTO Train/Test Graph.} In the common mode, we learn the total effect. Based on the causal graph in Fig.\ref{fig:3} (a), we intentionally add the links $X_{R} \rightarrow Y$ and $X_{T} \rightarrow Y$ to estimate the uni-modal direct effect of $X_{R}$ on $Y$ and $X_{T}$ on $Y$. We add these direct links for two reasons. 1) To estimate the input types as we will describe in Section 4.2. 2) To provide an estimate of the direct effect of $X_{T} \rightarrow Y$ in the differential mode. To implement this direct link, we train a uni-modal neural model denoted as $H_{\theta_{X_{R}}}(\cdot)$ and $H_{\theta_{X_{T}}}(\cdot)$. We denote the prediction scores obtained from direct links as $Y_{x_{{R}}}^{CM}$ and $Y_{x_{T}}^{CM}$ which can be expressed as below:
\begin{equation}{
Y_{m}^{CM}=H_{\theta_{M}}(m), Y_{x_{R}}^{CM}=H_{\theta_{X_{R}}}(x_{R}), Y_{x_{T}}^{CM}=H_{\theta_{X_{T}}}(x_{T}),
\label{eq:3}
}\end{equation}
where the neural network $H_{\theta_{M}}(\cdot)$ refers to the head network, which takes fusion features $M=m$ and outputs the class scores $Y_{m}^{CM}$ (direct link $M \rightarrow Y$). The structural causal model of \textit{common mode} can be illustrated as in Fig.\ref{fig:4} (a). \\
\indent In order to calculate prediction scores out of $Y_{m}^{CM}$, $Y_{x_{R}}^{CM}$, and $Y_{x_{T}}^{CM}$, we design a fusion function $\mathcal{F}(\cdot)$ using the nonlinear Log-Harmonic (LH):
\begin{equation}{
Y_{m, x_{R}, x_{T}}^{CM}=Y_{LH}^{CM}=log(\sigma(Y_{m}^{CM})\times\sigma(Y_{x_{R}}^{CM})\times\sigma(Y_{x_{T}}^{CM})),
\label{eq:4}
}\end{equation}
where $Y_{m, x_{R}, x_{T}}^{CM}$ denotes the final prediction score, and $\sigma$ denotes the sigmoid function. \\
\textbf{Total Effect (TE).} We measure the total effect by comparing $Y_{m, x_{R}, x_{T}}^{CM}$ with its no-treatment condition. Following the definition from the preliminaries, the total effect of the common mode causal graph can be written as:
\begin{equation}{
TE=Y_{m, x_{R}, x_{T}}^{CM}-Y_{m*, x_{R}*, x_{T}*}^{CM},
\label{eq:5}
}\end{equation}
where $Y_{m*, x_{R}*, x_{T}*}^{CM}$ is the no-treatment condition. The implementation details of the no-treatment condition are described in the \textit{supplementary material}. \\
\textbf{2. Differential Mode - RXTO Train/RXTO, ROTX Test Graph.} In the \textit{differential mode}, we learn the total indirect effect. Specifically, we intentionally assign node values and prune the direct link $X_{T}$ $\textcolor{red}{\rightarrow}$ $Y$. To this end, we introduce the natural direct effect.\\
\textbf{Natural Direct Effect (NDE).} We estimate the natural direct effect (NDE), the ``counterfactual scenario", which refers explicitly to the thermal direct effect i.e., $X_{T}$ $\textcolor{red}{\rightarrow}$ $Y$ in Fig.\ref{fig:4} (b). This direct effect can be estimated by blocking the effects of $X_{R}$ and $M$, using the no-treatment condition definitions in the preliminary section. $X_{T}$ is set to $x_{T}$, and $M$ would attain the value $m*$ when $X_{T}$ had been $x_{T}*$ and $X_{R}$ is $x_{R}*$. The natural direct effect (NDE) can be obtained by comparing the counterfactual graph to the no-treatment conditions:
\begin{equation}{
NDE=Y_{m*,x_{R}*,x_{T}}^{d}-Y_{m*,x_{R}*,x_{T}*}^{d},
\label{eq:6}
}\end{equation}
where $m*$, $x_{R}*$ and $x_{T}*$ represent the no-treatment condition.\\
\textbf{Total Indirect Effect (TIE).} The elimination of the thermal direct effect can be achieved by subtracting the natural direct effect (NDE) from the total effect (TE).
\begin{equation}{
TIE=TE-NDE=Y_{m, x_{R}, x_{T}}^{CM}-Y_{m*,x_{R}*, x_{T}}^{d}.
\label{eq:7}
}\end{equation}
Fig.\ref{fig:4} (c) shows an illustration of the total indirect effect of the differential mode. Note that the no-treatment conditions $Y_{m*,x_{R}*, x_{T}*}^{d}$ and $Y_{m*,x_{R}*, x_{T}*}^{CM}$ are the same. For inference, we opt for the label with the highest TIE, in contrast to the conventional strategies that primarily rely on posterior probability i.e., $p(y|x_{R},x_{T})$. 
\subsection{Causal Mode Multiplexing (CMM) Loss} 
\textbf{Determination of Causal Modes}. To assign different learning schemes based on the input type, we leverage the uni-modal prediction scores $Y_{x_{R}}^{CM}$ and $Y_{x_{T}}^{CM}$ based on the following rationale: For ROTO inputs, both the $Y_{x_{R}}^{CM}$ and $Y_{x_{T}}^{CM}$ value differences between pedestrians and backgrounds will have the same sign. In this case, we determine the input as ROTO and assign the common mode learning scheme. In the other case, we determine the input as RXTO or ROTX, assign the differential mode learning scheme, and prune the thermal direct link. Using these properties, we design a binary translation function to calculate the mode number. \\
\indent We denote the mode number as $K_{mode}$, for which we want the value 1 for the common mode and -1 for the differential mode. The details of our function design are as follows: Let $\pi_{R}=[\pi_{R}^{b},\pi_{R}^{f}$] and $\pi_{T}=[\pi_{T}^{b},\pi_{T}^{f}$] the approximate one-hot representation of the prediction labels derived from the $argmax$ of $Y_{x_{R}}^{CM}$ and $Y_{x_{T}}^{CM}$. Here, index 0 refers to the prediction label of the background (no pedestrian), and index 1 refers to the prediction label of the pedestrian. Since the $argmax$ operation is non-differentiable on the gradient descent method, we adopt the Gumbel-softmax~\citep{jang2016categorical} estimation. Then we can write $\pi_{R}$ and $\pi_{T}$ as:
\begin{equation}{
\pi_{R}=\textrm{softmax}\left[g_{R}+log(Y_{x_{R}}^{CM})/\tau\right], 
\label{eq:8}
}\end{equation}
\begin{equation}{
 \pi_{T}=\textrm{softmax}\left[g_{T}+log(Y_{x_{T}}^{CM})/\tau\right],
\label{eq:9}
}\end{equation}
where we set the Gumbel noises $g_{R}$ and $g_{T}$ to zero since we do not need random sampling variations for our purpose. From the formula of Eq.(\ref{eq:8}) and Eq.(\ref{eq:9}), we can obtain values $\left \{ \pi_{R}^{b}=0, \pi_{R}^{f}=1, \pi_{T}^{b}=0, \pi_{T}^{f}=1 \right \}$ for ROTO, $\left \{\pi_{R}^{b}=1, \pi_{R}^{f}=0, \pi_{T}^{b}=0, \pi_{T}^{f}=1\right \}$ for RXTO, and $\left \{\pi_{R}^{b}=0, \pi_{R}^{f}=1, \pi_{T}^{b}=1, \pi_{T}^{f}=0\right \}$ for ROTX. From these values, we can determine the causal mode number $K_{mode}$ according to the above rationale.
\begin{equation}{
\begin{aligned}
K_{mode}&=\Delta\pi_{R}\times\Delta\pi_{T}=(\pi_{R}^{f}-\pi_{R}^{b})\times(\pi_{T}^{f}-\pi_{T}^{b}).\\
\end{aligned}
\label{eq:10}
}\end{equation}
This $K_{mode}$ value will obtain 1 for the common mode and -1 for the differential mode. Then we can design a ``switchable total indirect effect" (sTIE) in which the cause-effect is differently calculated between TE and TIE according to the causal mode number $K_{mode}$.
\begin{equation}{
\begin{aligned}
sTIE&=TE-ReLU(-K_{mode})\times NDE \\
&=Y_{m, x_{R}, x_{T}}^{CM}-ReLU(-K_{mode})\times Y_{m*,x_{R}*, x_{T}}^{d}\\
&=\left\{\begin{matrix}
Y_{m, x_{R}, x_{T}}^{CM} \ \ \ \ \ \ \ \ \ \ \ \ \ \ \ \ \ \ \ \ \ \ \ \ if \ K_{mode}=1  \\ 
Y_{m, x_{R}, x_{T}}^{CM}-Y_{m*,x_{R}*, x_{T}}^{d} \ \ \ if \ K_{mode}=-1,
\end{matrix}\right.
\end{aligned}
\label{eq:11}
}\end{equation}

\begin{table}[t]
\centering
\caption{Cause-effect (CE) and $K_{mode}$ assignment with respect to data type.}
\resizebox{0.7\linewidth}{!}{
\begin{tabular}{c|c|c|c|c}
\Xhline{3\arrayrulewidth}
Type & $\Delta\pi_{R}$   & $\Delta\pi_{T}$   &  $K_{mode}$  & CE \\ \hline
ROTO & 1  & 1  & 1  & TE  \\ 
RXTO & -1 & 1  & -1 & TIE \\ 
ROTX & 1  & -1 & -1 & TIE \\ \Xhline{3\arrayrulewidth}
\end{tabular} 
}
\label{tab:1}
\end{table}

\noindent in which sTIE is calculated as the total effect in Eq.(\ref{eq:5}) for ROTO inputs and as the total indirect effect in Eq.(\ref{eq:7}) for RXTO and ROTX inputs. We summarize the relationship between cause-effect (CE) and $K_{mode}$ assignment with respect to data type in Table \ref{tab:1}.\\
\textbf{Causal Mode Multiplexing Loss}.
We formulate the sTIE as a loss function. Before introducing our Causal Mode Multiplexing Loss, we first revisit the loss function of the conventional model. Given a triplet $(x_{T}, x_{R}, y)$ where $y$ is the ground-truth class label of RGB/T ROI feature pair: $x_{R}$/$x_{T}$, the ROI classification branches of the conventional multispectral pedestrian detection model~\citep{9419080} are optimized by:
\begin{equation}{
\begin{aligned}
\ \ \ \ \ \mathcal{L}_{cls}=\mathcal{L}_{Y}(TE,y) \ \  \textbf{\textit{(Conventional)}},
\end{aligned}
\label{eq:12}
}\end{equation}
where $\mathcal{L}_{Y}$ denotes the cross-entropy loss. This conventional loss guides the model to learn the total effect on both ROTO and RXTO, which provokes the thermal direct effect. Different from them, Causal Mode Multiplexing (CMM) Loss learns the causality in both ROTO and RXTO training data based on two causal modes:
\begin{equation}{
\ \ \mathcal{L}_{CMM}=\mathcal{L}_{Y}(sTIE,y),
\label{eq:13}
}\end{equation}
which sTIE refers to the formula in Eq.(\ref{eq:11}). The overall classification branch loss can be written as:
\begin{equation}{
\ \ \mathcal{L}_{cls}=\mathcal{L}_{CMM}+\mathcal{L}_{Y}(Y_{x_{R}}^{CM},y)+\mathcal{L}_{Y}(Y_{x_{T}}^{CM},y),
\label{eq:14}
}\end{equation}
where $\mathcal{L}_{Y}(Y_{x_{R}}^{CM},y)$ and $\mathcal{L}_{Y}(Y_{x_{T}}^{CM},y)$ are over $Y_{x_{R}}^{CM}$ and $Y_{x_{T}}^{CM}$.
\subsection{Implementation} 
\textbf{Training}. The final training loss is the combination of $\mathcal{L}_{cls}$, bounding box regression loss $\mathcal{L}_{bbox}$, and $\mathcal{L}_{model}$, which include the RPN and the uncertainty module. We follow the implementation details of the paper~\citep{9419080}.
\begin{equation}{
\mathcal{L}_{total}=\sum_{(x_{R},x_{T},y)\in D}\mathcal{L}_{cls}+\mathcal{L}_{bbox}+\mathcal{L}_{model}.
\label{eq:15}
}\end{equation}
\textbf{Inference}. We use the switchable total indirect effect (sTIE) for inference.
\begin{equation}{
sTIE=Y_{m, x_{R}, x_{T}}^{CM}-ReLU(-K_{mode})\times Y_{m*,x_{R}*, x_{T}}^{d}.
}
\label{eq:16}
\end{equation}
\section{New Dataset: ROTX-MP}
\noindent \indent To evaluate modality bias in multispectral pedestrian detectors, we propose to collect a new dataset: The ROTX Multispectral Pedestrian (ROTX-MP) dataset. ROTX-MP consists of 1000 ROTX test images collected from two practical scenarios related to the applications of multispectral pedestrian detection. The details of ROTX-MP are described below.
\subsection{Comparison to Existing Datasets}
\noindent \indent Our ROTX-MP dataset mainly contains ROTX data, compared to existing datasets that consist of ROTO and RXTO data. We compare the data distributions of datasets using the statistics criteria as the following.
\subsubsection{Statistics Criteria}
\noindent \indent For the distribution comparison, we count the number of images for each data type: ROTO, RXTO, and ROTX. We explain our criteria for counting ROTO, RXTO, and ROTX (or ROTO/ROTX) data in datasets.\\
\textbf{ROTO}: We count the number of daytime images for each dataset for ROTO. \\
\textbf{RXTO}: We count the nighttime images for each dataset for RXTO.\\
\textbf{ROTX}: ROTX data refer to daytime images where there is an intermediate medium that obscures the thermal radiation of the pedestrian. In this case, pedestrians are only imaged by RGB sensors. We count images that contain these scenarios. There can be images where ROTO and ROTX scenarios both occur. We note them separately, as ROTO/ROTX. 
\subsubsection{Dataset Distribution}
\begin{figure}[t]
  \centering
\includegraphics[width=\linewidth]{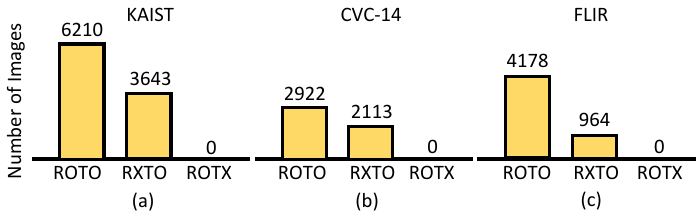}
  \caption{ROTO, RXTO, and ROTX distribution of popular multispectral pedestrian datasets: (a) KAIST~\cite{hwang2015multispectral}, (b) CVC-14~\cite{gonzalez2016pedestrian}, and (c) FLIR~\cite{c:25}. Images from all train/test sets are counted.}
    \label{fig:5}
    \vspace{-0.4cm}
\end{figure}

\begin{figure}[t]
  \centering
\includegraphics[width=0.9\linewidth]{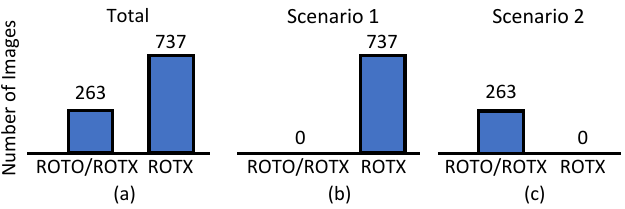}
  \caption{ROTO/ROTX and ROTX distribution of ROTX-MP test set.}
    \label{fig:6}
    \vspace{-0.5cm}
\end{figure}

\noindent \indent From the above criteria, we obtain the distribution of each ROTO, RXTO, and ROTX (or ROTO/ROTX) in different datasets: 1) KAIST~\cite{hwang2015multispectral}, 2) CVC-14~\cite{gonzalez2016pedestrian}, 3) FLIR~\cite{c:25}, and 4) our ROTX-MP. We count the images of all train and test data for KAIST, CVC-14, and FLIR, as well as the images of the test data of ROTX-MP. ROTX-MP contains only test data. The distribution is plotted in Fig.\ref{fig:5} and Fig.\ref{fig:6}. It can be seen from Fig.5 that KAIST, CVC-14, and FLIR mainly contain ROTO and RXTO data and no ROTX scenes (including ROTO/ROTX). On the other hand, the ROTX-MP dataset (1000 images) primarily consists of ROTX data (737 images) and ROTO/ROTX (263 images) data. Moreover, we compare the overall size of the dataset. Since ROTX-MP contains only test data, we compare it with the test sets of KAIST, CVC-14, and FLIR. ROTX-MP contains 1000 test data, comparable to KAIST test-2252 images (ROTO: 1455, RXTO: 797), CVC-14 test-1417 images (ROTO: 690, RXTO: 727), and FLIR test-1013 images (ROTO: 702, RXTO: 311). 
\subsection{Data Collection Process}
\noindent \indent ROTX-MP is a dataset of RGBT multispectral image pairs captured from an RGBT camera. The RGBT camera we used for data collection was the FLIR Duo Pro R (FPA $640\times512$) camera manufactured by FLIR Systems, Inc. This product supports simultaneous RGB and thermal ($NETD<50 mK$, $\lambda= 7.5\sim 13.5\mu m$) imaging in a dual pip mode. We invited 8 volunteers to participate in our data collection. All of the volunteers agreed to the data collection and its future release to the public. On streets and indoors, we mounted the camera device on a portable tripod and captured video of volunteers from different shooting distances ranging from 0 to 15 meters. Each video involved 2–8 volunteers. From the recorded videos, we meticulously hand-picked high-quality image pairs. RGBT pairs were spatially aligned and synchronized, and finalized of 1000 RGBT pairs. The faces of volunteers are blurred for privacy protection. Specifically, we collected ROTX data based on the below practical scenarios.\\
\hspace{-0.3cm}\textbf{1.\,Pedestrians Over a Glass Window (737 ROTX images).} Multispectral pedestrian detection is attractive for all-day/night smart surveillance. However, pedestrians observed through a window are visible in RGB but obscured in thermal because thermal radiation cannot penetrate through glass. Failure in these scenes constrains multispectral pedestrian detection within the window.  \\
\textbf{2.\,Pedestrians Wearing Heat-insulation Clothes (263 ROTO+ROTX images).} Heat-insulation clothing such as fire protection gear or low-emissivity clothing provides a way of thermal invisibility or \textit{stealth}. Wearing fire protection gear can make firefighters or evacuees undetectable from autopilots equipped with multispectral cameras in search and rescue (SAR) situations. Also, criminals (e.g., bank robbers) can wear low-emissivity clothing to evade multispectral pedestrian detection on security cameras. We collected 150 images for firefighter scenes and 113 images for criminal scenes. In both scenarios, we simultaneously contain ROTO scenes for comparison to the obscured pedestrians. Thus, we note the ROTO+ROTX data type for this scenario. 
\section{Experimental Setup}
\subsection{Dataset}
\noindent \indent We conduct experiments on multiple multispectral pedestrian datasets: KAIST~\cite{hwang2015multispectral}, CVC-14~\cite{gonzalez2016pedestrian}, FLIR~\cite{c:25}, and our collected dataset: ROTX-MP. KAIST contains 95,328 RGBT data with $640\times512$ image resolutions. We use the annotation labels provided by the original authors~\cite{hwang2015multispectral} and organize the dataset by 7601 training images and 2252 test images following the baselines~\cite{9419080,zhou2020improving} we compare. The 2252 (`All') test images are composed of 1455 (`Day’) images and 797 (`Night’) images. CVC-14~\cite{gonzalez2016pedestrian} contains 3618 training data and 1417 test data with grayscale RGBT images of $640\times471$ resolution. 1417 test data (`All') include 690 daytime images (`Day') and 727 nighttime images (`Night'). Also, we evaluate under the Teledyne FLIR Free ADAS Thermal Dataset v2.0.0~\cite{c:25} (FLIR for short). Following previous works~\cite{zhang2020multispectral, qingyun2022crossmodality, kim2023multispectral}, we use 4129 train image pairs and 1013 test image pairs organized by ~\citet{zhang2020multispectral} for fair comparison. As evaluations in previous work do not compare day and night performance separately, we note (`All') for the FLIR testset. ROTX-MP is composed of 1000 test images. Similar to FLIR, we note (`All') for ROTX-MP.

\subsection{Baseline Model}
\noindent \indent We implement CMM based on the Uncertainty-Guided model~\cite{9419080}. Our method is compared with four competitive multimodal pedestrian detection architectures recently proposed: Kim et al.~\citep{9419080}, Cross-modality Fusion Transformer (CFT)~\citep{qingyun2022crossmodality}, MBNet~\citep{zhou2020improving}, and Halfway Fusion~\citep{liu2016multispectral}+Faster RCNN~\cite{ren2015faster} version (HFF). More details are described in the supplementary material.
\section{Experimental Results}
\begin{figure*}[t!]
  \centering
  \includegraphics[width=0.88\linewidth]{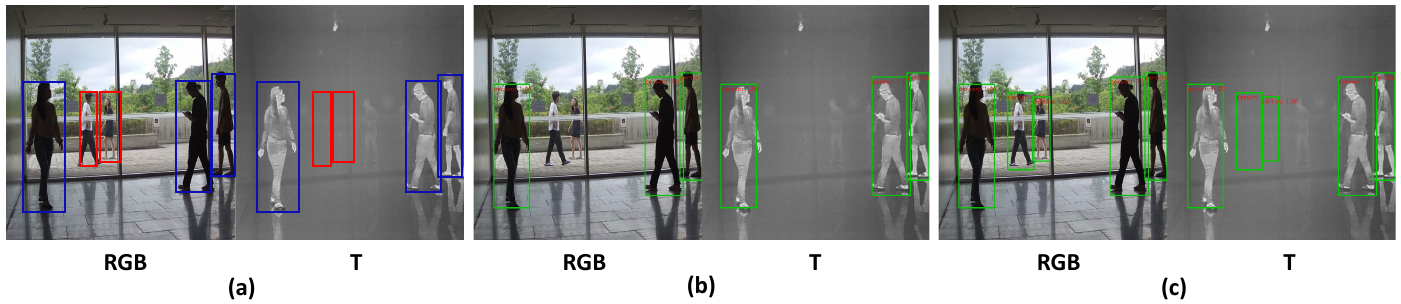}
  \caption{\textbf{Visualized examples of multispectral pedestrian detection on ROTO/ROTX data}. (a) \textcolor{blue}{ROTO} and \textcolor{red}{ROTX} scenarios. (b) Conventional models~\cite{9419080} fail to detect pedestrians in ROTX scenes. (c) Our CMM framework estimates the total effect for ROTO and the total indirect effect for ROTX. As a result, it produces correct debiased results for ROTX samples, even from biased training data. We provide more examples in the supplementary material.}
    \label{fig:7}
\end{figure*}
\begin{table}[t!]
\caption{Detection performance of different multispectral pedestrian detection models on the KAIST, CVC-14, and FLIR datasets.}
	\renewcommand{\arraystretch}{1.2}
\centering
\resizebox{0.8\linewidth}{!}{
\begin{tabular}{cl|ccc|ccc|c}
\Xhline{3\arrayrulewidth}
\multicolumn{2}{c|}{Dataset}    & \multicolumn{3}{c|}{KAIST}                    & \multicolumn{3}{c|}{CVC-14}                      & FLIR           \\ \hline
\multicolumn{2}{c|}{Metric}     & \multicolumn{3}{c|}{MR$(\downarrow)$}                       & \multicolumn{3}{c|}{MR$(\downarrow)$}                          & AP$(\uparrow)$             \\ \hline
\multicolumn{2}{c|}{Model}      & Day           & Night         & All           & Day            & Night          & All            & All            \\ \hline
\multicolumn{2}{c|}{HFF}        & 14.24         & 10.26         & 13.04         & 50.48          & 36.05          & 44.31          & 75.85          \\
\multicolumn{2}{c|}{CFT}        & 13.99         & 7.05          & 11.60         & \textbf{18.81}          & 25.25          & 21.83          & 84.10          \\
\multicolumn{2}{c|}{MBNet}      & 9.64 & 8.26          & 9.04          & 24.70 & 13.50          & 21.10          & -              \\
\multicolumn{2}{c|}{Kim et al.} & 10.11         & \textbf{5.05} & 8.67 & 23.87 & 11.08 & 18.70 & 84.67 \\ \hdashline
\multicolumn{2}{c|}{CMM (Ours)} & \textbf{9.60} & 5.93 & \textbf{8.54} & 27.81          & \textbf{7.71}  & \textbf{17.13} & \textbf{87.80} \\ \Xhline{3\arrayrulewidth}
\end{tabular}}
\label{tab:2}
\vspace{-0.3cm}
\end{table}
\subsection{Evaluation on Existing Datasets}
\noindent \indent We first report the experimental results on the KAIST
~\cite{hwang2015multispectral}, CVC-14~\cite{gonzalez2016pedestrian}, and FLIR~\cite{c:25} data. These datasets are comprised of day (ROTO) and night (RXTO) images. We train and test on these datasets to evaluate whether CMM over-corrects the modality bias. Our motivation for designing the CMM framework is to learn causality based on different input types (ROTO, and RXTO) in the training set; therefore, CMM's intentional behavior is to generalize on existing data as well. \\
\indent For the evaluation metrics, we follow previous multispectral pedestrian detection works for measuring performance. Specifically, we calculate the Log Average Miss-Rate (MR$\downarrow$) for the `All', `Day', and `Night' test images of the KAIST and CVC-14 test sets. From our definition, `Day' can be interpreted as ROTO data, `Night' as RXTO data, and `All' as their combination. To evaluate performance on the FLIR dataset, we use the Average Precision (AP$\uparrow$) to conduct a fair comparison with other methods~\cite{qingyun2022crossmodality}. A low Miss Rate and a high Average Precision value indicate high detection performance. \\
\indent Results are reported in Table \ref{tab:2}. The CMM framework demonstrates the lowest miss rates for `Day' and `All' in the KAIST set, `Night' and `All' in the CVC-14 set, and the highest AP in the FLIR set. Since `All' represents the full dataset, it can be concluded that CMM achieves the best performance across all KAIST, CVC-14, and FLIR datasets. It is worth noting that CMM performs better than its baseline model, ~\cite{9419080}, with 0.13 MR, 1.57 MR, and 3.13 AP improvements in KAIST, CVC-14, and FLIR, respectively. Such results meet the original intention of our CMM design, and we extend the evaluation on ROTX-MP.
\subsection{Evaluation on ROTX-MP}
\noindent \indent The CMM framework is introduced to enhance the generalizability of models when there is a substantial distribution difference in the training and test splits. Therefore, we train models on conventional datasets~\cite{hwang2015multispectral, gonzalez2016pedestrian,c:25} and test on ROTX-MP to evaluate the generalizability on ROTX data of each model. Here, we do not directly compare with data augmentation methods~\cite{chang2021towards,agarwal2020towards}, which explicitly generate training samples, e.g., counterfactual generation. Generalizing on ROTX data by modifying training priors violates the original purpose of the CMM model, i.e., to evaluate whether multispectral pedestrian detection models infer from memorized priors in training data.\\
\indent For each evaluation on the ROTX-MP test set (1000 test images), we calculate the Average Precision (AP $\uparrow$) since it is a standard metric to analyze the accuracy of a detection model. Evaluations are reported in Table \ref{tab:4}. Other models fail on ROTX-MP test data when trained on conventional datasets. Models trained on the KAIST~\cite{hwang2015multispectral} dataset: HFF~\cite{liu2016multispectral}, CFT~\cite{qingyun2022crossmodality}, MBNET~\cite{zhou2020improving}, ~\citet{9419080} achieve 36.95, 3.64, 18.88, 21.69 AP on ROTX-MP. Compared to these methods, CMM shows superior generalizability on ROTX-MP test data, achieving 70.44 AP, outperforming other baselines with at least 33.49 AP. Similar results are obtained from models trained on CVC-14 and FLIR when tested on ROTX-MP. Fig. \ref{fig:7} shows qualitative results. 
\begin{table*}[t!]
\caption{\textbf{Ablation study on debiasing strategy.} ``A+B” denotes the debiasing strategy that train the model with ``A" cause-effect and test with ``B" cause-effect. The baseline, TE+TIE, TIE+TIE, and sTIE (ours) are compared. }
	\renewcommand{\arraystretch}{1.8}
	\renewcommand{\tabcolsep}{1mm}
\centering
\resizebox{0.8\linewidth}{!}{
\begin{tabular}{c|cccccc|cccccc|cc}
\clineB{1-5}{4} \clineB{7-11}{4} \clineB{13-15}{4}
Train    & \multicolumn{4}{c}{KAIST}                                                           &  & Train    & \multicolumn{4}{c}{CVC-14}                                                            &  & Train    & \multicolumn{2}{c}{FLIR}                             \\ \cline{1-5} \cline{7-11} \cline{13-15} 
Test     & \multicolumn{1}{c|}{ROTX-MP (AP$\uparrow$)} & \multicolumn{3}{c}{KAIST (MR$\downarrow$)}                &  & Test     & \multicolumn{1}{c|}{ROTX-MP (AP$\uparrow$)} & \multicolumn{3}{c}{CVC-14 (MR$\downarrow$)}                 &  & Test     & \multicolumn{1}{c|}{ROTX-MP (AP$\uparrow$)} & FLIR (AP$\uparrow$)      \\ \cline{1-5} \cline{7-11} \cline{13-15} 
Model    & \multicolumn{1}{c|}{All}            & Day           & Night         & All           &  & Model    & \multicolumn{1}{c|}{All}            & Day            & Night         & All            &  & Model    & \multicolumn{1}{c|}{All}            & All            \\ \cline{1-5} \cline{7-11} \cline{13-15} 
Baseline & \multicolumn{1}{c|}{21.69}          & 10.11         & \textbf{5.05} & 8.67          &  & Baseline & \multicolumn{1}{c|}{13.36}          & 23.87 & 11.08         & 18.70          &  & Baseline & \multicolumn{1}{c|}{12.23}          & 84.67          \\
TE+TIE   & \multicolumn{1}{c|}{57.05}          & 26.89         & 27.66         & 27.27         & \hspace{0.5cm}  & TE+TIE   & \multicolumn{1}{c|}{27.97}          & 32.85          & 12.11         & 22.33          &  & TE+TIE   & \multicolumn{1}{c|}{19.33}          & 64.27          \\
TIE+TIE  & \multicolumn{1}{c|}{56.45}          & 12.53         & 9.24          & 12.10         &  & TIE+TIE  & \multicolumn{1}{c|}{27.75}          & 33.53          & 9.27          & 21.63          & \hspace{0.5cm} & TIE+TIE  & \multicolumn{1}{c|}{12.02}          & 79.39          \\
\cdashline{1-5} \cdashline{7-11} \cdashline{13-15}
sTIE (Ours)      & \multicolumn{1}{c|}{\textbf{70.44}} & \textbf{9.60} & 5.93          & \textbf{8.54} &  & sTIE (Ours)      & \multicolumn{1}{c|}{\textbf{34.96}} & 27.81          & \textbf{7.71} & \textbf{17.13} &  & sTIE (Ours)      & \multicolumn{1}{c|}{\textbf{57.09}} & \textbf{87.80} \\ \clineB{1-5}{4} \clineB{7-11}{4} \clineB{13-15}{4} 
\end{tabular}}
\label{tab:3}
\end{table*}
\vspace{-0.3cm}
\section{Ablation Study}
\noindent \indent Furthermore, we conduct ablation studies to validate the design choice of the switchable total indirect effect (sTIE) in Eq.(\ref{eq:11}) for our debiasing strategy. Known debiasing strategies~\cite{niu2021counterfactual, tian2022debiasing} are generally based on training the model via total effect (TE) and inferring with total indirect effect (TIE). For the comparison study, we conducted an experiment on (2) TE training + TIE inference of the (1) baseline model~\cite{9419080}. Moreover, we also evaluate the (3) TIE training + TIE inference strategy to provide an extensive comparison. The results are shown in Table \ref{tab:3}. Firstly, the (1) baseline model performs well on existing datasets but generalizes poorly on ROTX-MP. Secondly, the (2) TE+TIE model demonstrates enhanced model generalizability for ROTX data compared to the baseline, but degrades on existing data. Moreover, the (3) TIE+TIE model demonstrates a moderate level of performance, yet degrades on existing datasets compared to the baseline. Compared to them, the (4) CMM framework shows superior performance on both ROTX-MP and conventional test sets.  From this ablation study, we verify the effectiveness of our design choice of the switchable total indirect effect (sTIE) in Eq.(\ref{eq:11}).
\begin{table}[t!]
\caption{Detection performance on the ROTX-MP test set. Models are trained from the KAIST, CVC-14, and FLIR datasets.}
	\renewcommand{\arraystretch}{1.2}
\centering
\resizebox{0.65\linewidth}{!}{

\begin{tabular}{cl|c|c|c}
\Xhline{3\arrayrulewidth}
\multicolumn{2}{c|}{Train}      & KAIST                & CVC-14               & FLIR                 \\ \hline
\multicolumn{2}{c|}{Test}       & ROTX-MP            & ROTX-MP            & ROTX-MP            \\ \hline
\multicolumn{2}{c|}{Metric}     & AP$(\uparrow)$                   & AP$(\uparrow)$                   & AP$(\uparrow)$                   \\ \hline
\multicolumn{2}{c|}{Model}      & All                  & All                  & All                  \\ \hline
\multicolumn{2}{c|}{HFF}        & 36.95       & 8.80                 & 13.21       \\
\multicolumn{2}{c|}{CFT}        & 3.64                 & 8.58                 & 5.28                 \\
\multicolumn{2}{c|}{MBNet}      & 18.88                & -                    & -                    \\
\multicolumn{2}{c|}{Kim et al.} & 21.69                & 13.36       & 12.23                \\ \hdashline
\multicolumn{2}{c|}{CMM (Ours)} & {\textbf{70.44}}& {\textbf{34.96}} & {\textbf{57.09}} \\ \Xhline{3\arrayrulewidth}
\end{tabular}
}
\label{tab:4}
\vspace{-0.6cm}
\end{table}
\vspace{-0.2cm}
\section{Conclusion}
\noindent \indent In this paper, we address the modality bias problem in multispectral pedestrian detection with our innovative solution: The Causal Mode Multiplexer (CMM) framework. Using the tools of counterfactual intervention, CMM enables the model to interchangeably learn between two distinct causal graphs depending on the input data type. We propose the Causal Mode Multiplexing (CMM) Loss to optimize the interchange between two causal graphs. Additionally, we introduce the ROTX-MP dataset to evaluate modality bias in multispectral pedestrian detection. Experimental results on KAIST, CVC-14, FLIR, and our ROTX-MP dataset demonstrate that CMM effectively learns multimodal reasoning and performs well on ROTX test data with training ROTO and RXTO data.
\vspace{-0.2cm}
\section{Acknowledgements}
This work was partly supported by two
funds: IITP grant funded by the Korea government (MSIT)
(No.2022-0-00984) and Center for Applied Research in Artificial Intelligence (CARAI) grant funded by DAPA and
ADD (UD230017TD).
{
    \small
    \bibliographystyle{ieeenat_fullname}
    \bibliography{main}
}


\end{document}